\documentclass[11pt,letterpaper]{article}
\usepackage{naaclhlt2016}
\usepackage{times}
\usepackage{latexsym}
\usepackage{amsmath}
\usepackage{amssymb}
\usepackage{graphicx}
\usepackage{subcaption}
\usepackage{url}

\naaclfinalcopy 
\def\naaclpaperid{38} 

\title{What is the Role of Recurrent Neural Networks (RNNs) in an Image Caption Generator?}

\author{Marc Tanti \qquad Albert Gatt\\
	    Institute of Linguistics \\and Language Technology\\
	    University of Malta\\
	    {\tt marc.tanti.06@um.edu.mt}\\
        {\tt albert.gatt@um.edu.mt}        
	  	\And
		Kenneth P. Camilleri\\
  		Deptartment of Systems \\and Control Engineering\\
		University of Malta\\
  		{\tt kenneth.camilleri@um.edu.mt}
  }

\date{May 2017}

\begin{document}

\maketitle

\begin{abstract}
In neural image captioning systems, a recurrent neural network (RNN) is typically viewed as the primary `generation' component. This view suggests that the image features should be `injected' into the RNN. This is in fact the dominant view in the literature. Alternatively, the RNN can instead be viewed as only encoding the previously generated words. This view suggests that the RNN should only be used to encode linguistic features and that only the final representation should be `merged' with the image features at a later stage. This paper compares these two architectures. We find that, in general, late merging outperforms injection, suggesting that RNNs are better viewed as encoders, rather than generators.
\end{abstract}

\section{Introduction}\label{sec:intro}
Image captioning \cite{Bernardi2016} has emerged as an important testbed for solutions to the fundamental AI challenge of grounding symbolic or linguistic information in perceptual data \cite{Harnad1990,Roy2005}. Most captioning systems focus on what \newcite{Hodosh2013} refer to as {\em concrete conceptual} descriptions, that is, captions that describe what is strictly within the image, although recently, there has been growing interest in moving beyond this, with research on visual question-answering \cite{Antol2015} and image-grounded narrative generation \cite{Huang2016} among others. 

Approaches to image captioning can be divided into three main classes \cite{Bernardi2016}:

\begin{enumerate}
\item Systems that rely on computer vision techniques to extract object detections and features from the source image, using these as input to an NLG stage \cite{Kulkarni2011,Mitchell2012,Elliott2013}. The latter is roughly akin to the microplanning and realisation modules in the well-known NLG pipeline architecture  \cite{Reiter2000}.
\item Systems that frame the task as a retrieval problem, where a caption, or parts thereof, is identified by computing the proximity/relevance of strings in the training data to a given image. This is done by exploiting either a unimodal \cite{Ordonez2011,Gupta2012,Mason2014} or multimodal \cite{Hodosh2013,Socher2014} space. Many retrieval-based approaches rely on neural models to handle both image features and linguistic information \cite{Ordonez2011,Socher2014}.
\item Systems that also rely on neural models, but rather than performing partial or wholesale caption retrieval, generate novel captions using a recurrent neural network (RNN), usually a long short-term memory (LSTM). Typically, such models use image features extracted from a pre-trained convolutional neural network (CNN) such as the VGG CNN \cite{Simonyan2014} to bias the RNN towards sampling terms from the vocabulary in such a way that a sequence of such terms produces a caption that is relevant to the image \cite{Kiros2014,Kiros2014a,Vinyals2015,Mao2015,Hendricks2016}.
\end{enumerate}

This paper focuses on the third class. The key property of these models is that the CNN image features are used to condition the predictions of the best caption to describe the image. However, this can be done in different ways and the role of the RNN depends in large measure on the mode in which CNN and RNN are combined.

It is quite typical for RNNs to be viewed as `generators'. For example, \newcite{Bernardi2016} suggest that `the RNN is trained to generate the next word [of a caption]', a view also expressed by \newcite{LeCun2015}. A similar position has also been taken in work focusing on the use of RNNs as language models for generation \cite{Sutskever2011,Graves2013}. 
However, an alternative view is possible, whereby the role of the RNN can be thought of as primarily to encode sequences, but not directly to generate them.

\begin{figure}[!h]
	\centering
	\begin{subfigure}{0.45\textwidth}
      \caption{
          \label{fig:conditioning_inject}
          Conditioning by injecting the image means injecting the image into the same RNN that processes the words.
      }
      \centering
      \includegraphics[width=0.7\textwidth]{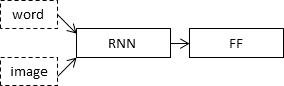}
	\end{subfigure}
	
    \vspace{10pt}
    
	\begin{subfigure}{0.45\textwidth}
      \caption{
          \label{fig:conditioning_merge}
          Conditioning by merging the image means merging the image with the final state of the RNN in a ``multimodal layer'' after processing the words.
      }
      \centering
      \includegraphics[width=0.7\textwidth]{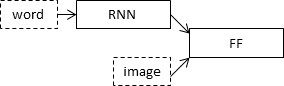}
    \end{subfigure}
	\caption{
		\label{fig:conditioning}
		The inject and merge architectures for caption generation. The RNN's previous state going into the RNN is not shown. Legend: RNN - Recurrent Neural Network; FF - Feed Forward layer.
	}
\end{figure}

These two views can be associated with different architectures for neural caption generators, which we discuss below and illustrated in Figure~\ref{fig:conditioning}. In one class of architectures, image features are directly incorporated into the RNN during the sequence encoding process (Figure~\ref{fig:conditioning_inject}). In these models, it is natural to think of the RNN as the primary generation component of the image captioning system, making predictions conditioned by the image. A different architecture keeps the encoding of linguistic and perceptual features separate, merging them in a later multimodal layer, at which point predictions are made (Figure~\ref{fig:conditioning_merge}). In this type of model, the RNN is functioning primarily as an encoder of sequences of word embeddings, with the visual features merged with the linguistic features in a later, multimodal layer. This multimodal layer is the one that drives the generation process since the RNN never sees the image and hence would not be able to direct the generation process.

While both architectural alternatives have been attested in the literature, their implications have not, to our knowledge, been systematically discussed and comparatively evaluated. In what follows, we first  discuss the distinction between the two architectures (Section \ref{sec:background}) and then present some experiments comparing the two (Sections \ref{sec:experiments} and \ref{sec:results}). Our conclusion is that grounding language generation in image data is best conducted in an architecture that first encodes the two modalities separately, before merging them to predict captions.

\section{Background: Neural Caption Generation Architectures}\label{sec:background}

\begin{figure}[t]
	\centering
	  \includegraphics[width=0.45\textwidth]{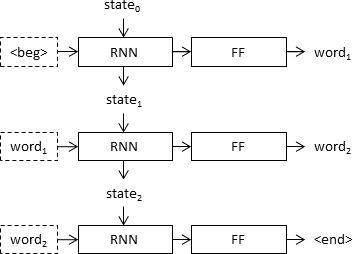}	
      \caption{
          \label{fig:rnn-cont}
          How RNNs work: each state of the RNN encodes a prefix, which incorporates the output word derived from the previous state. In practice the neural network does not output a single word but a probability distribution over all known words in the vocabulary. Legend: FF - feedforward layer; \textless beg\textgreater\ - the start-of-sentence token; \textless end\textgreater\ - the end-of-sentence token. 
      }
\end{figure}

In a neural language model, an RNN encodes a prefix (for example, the caption generated so far) and either itself predicts the next item in the sequence with the help of a feed forward layer or else it passes the encoding to the next layer which will make the prediction itself. This new item is added to the prefix at the next iteration to predict another item, until an end-of-sequence symbol is reached. Typically, the prediction is carried out using a softmax function to sample the next item according to a probability distribution over the vocabulary items, based on their activation. This process is illustrated in Figure \ref{fig:rnn-cont}.

One way to condition the RNN to predict image captions is to inject both visual and linguistic features directly into the RNN, depicted in Figure \ref{fig:conditioning_inject}. We refer to this as `conditioning-by-inject' (or {\em inject} for short). Different types of inject architectures have become the most widely attested among deep learning approaches to image captioning \cite{Chen2015,Donahue2015,Hessel2015,Karpathy2015,Liu2016,Yang2016,Zhou2016}.\footnote{See \newcite{TantiGC17} for an overview of different versions of the inject architecture and a systematic comparison among models. In this paper we focus on parallel-inject.} Given training pairs consisting of an image and a caption, the RNN component of such models is trained by exposure to prefixes of increasing length extracted from the caption, in tandem with the image.

An alternative architecture -- which we refer to as `conditioning-by-merge' (Figure \ref{fig:conditioning_merge}) -- treats the RNN exclusively as a `language model' to encode linguistic sequences of varying length. The linguistic vector resulting from this encoding is subsequently combined with the image features in a separate multimodal layer. This amounts to viewing the RNN as primarily an encoder of linguistic information.
This type of architecture is also attested in the literature, albeit to a lesser extent than the inject architecture 
\cite{Mao2014,Mao2015,Mao2015a,Song2016,Hendricks2016,You2016}. 
A limited number of approaches have also been proposed in which both architectures are combined \cite{Lu2016,Xu2015}.

Notice that both architectures are compatible with the inclusion of attentional mechanisms \cite{Xu2015}. The effect of attention in the inject architecture is to combine a different representation of the image with each word. In the case of merge, a different representation of the image can be combined with the final RNN state before each prediction. Attentional mechanisms are however beyond the scope of the present work.

The main differences between inject and merge architectures can be summed up as follows: In an inject model, the RNN is trained to predict sequences based on histories consisting of both linguistic and perceptual features. Hence, in this model, the RNN is primarily responsible for image-conditioned language generation. By contrast, in the merge architecture, RNNs in effect encode linguistic {\em representations}, which themselves constitute the input to a later prediction stage that comes after a multimodal layer. It is only at this late stage that image features are used to condition predictions.

As a result, a model involving conditioning by inject is trained to learn linguistic representations directly conditioned by image data; a merge architecture maintains a distinction between the two representations, but brings them together in a later layer.

Put somewhat differently, it could be argued that at a given time step, the merge architecture predicts what to generate next by combining the RNN-encoded prefix of the string generated so far (the `past' of the generation process) with non-linguistic information (the guide of the generation process). The inject architecture on the other hand uses the full image features with every word of the prefix during training, in effect learning a `visuo-linguistic' representation of each word. One effect of this is that image features can serve to further specify or disambiguate the `meaning' of words, by disambiguating tokens of the same word which are correlated with different image features (such as `crane' as in the bird versus the construction equipment). This implies that inject models learn a larger vocabulary during training.

The two architectures also differ in the number of parameters they need to handle. As noted above, since an inject architecture combines the image with each word during training, it is effectively handling a larger vocabulary than merge. Assume that the image vectors are concatenated with the word embedding vectors (inject) or the final RNN state (merge). Then, in the inject architecture, the number of weights in the RNN is a function of both the caption embedding and the images, whereas in merge, it is only the word embeddings that contribute to the size of this layer of the network. Let $e$ be the size of the word embedding, $v$ the size of the vocabulary, $i$ the image vector size and $s$ the state size of the RNN. In the inject case, the number of weights in the RNN is $w \propto (e+i) \times s$, whereas it is $w \propto e \times s$ in merge. The smaller number of weights handled by the RNN in merge is offset by a larger number of weights at the final softmax layer, which has to take as input the RNN state and the image, having size $\propto (s + i) \times v$.

A systematic comparison of these two architectures would shed light on the best way to conceive of the role of RNNs in neural language generation. Apart from the theoretical implications concerning the stage at which language should be grounded in visual information, such a comparison also has practical implications. In particular, if it turns out that merge outperforms inject, this would imply that the linguistic representations encoded in an RNN could be pre-trained and re-used for a variety of tasks and/or image captioning datasets, with domain-specific training only required for the final feedforward layer, where the tuning required to make perceptually grounded predictions is carried out. We return to this point in Section \ref{sec:future}.

In the following sections, we describe some experiments to conduct such a comparison.

\section{Experiments}\label{sec:experiments}

\begin{figure}
	\centering
	
	\begin{subfigure}{0.45\textwidth}
      \caption{
          \label{fig:architectures_merge}
          The merge architecture.
      }
      \centering
      \includegraphics[width=\textwidth]{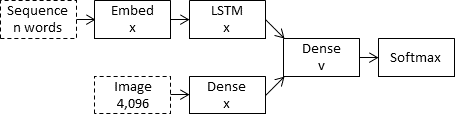}
	\end{subfigure}
    
	\vspace{10pt}
	
    \begin{subfigure}{0.45\textwidth}
      \caption{
          \label{fig:architectures_inject}
          The inject architecture.
      }
      \centering
      \includegraphics[width=\textwidth]{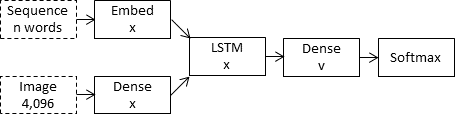}
	\end{subfigure}
	
	\caption{
		\label{fig:architectures}
		An illustration of the different architectures that are tested in this paper. The numbers or letters at the bottom of each box refer to the vector size output of a layer. `x' is  an arbitrary layer size that is varied in the experiments and `v' is the vocabulary size which is also varied in the experiments. `Dense' means fully connected layer with bias.
	}
\end{figure}

To evaluate the performance of the inject and merge architectures, and thus the roles of the RNN, we trained and evaluated them on the Flickr8k \cite{Hodosh2013} and Flickr30k \cite{Young2014} datasets of image-caption pairs. For the purposes of these experiments, we used the version of the datasets distributed by \newcite{Karpathy2015}\footnote{\url{http://cs.stanford.edu/people/karpathy/deepimagesent/}}. The dataset splits are identical to that used by \newcite{Karpathy2015}: Flickr8k is split into 6,000 images for training, 1,000 for validation, and 1,000 for testing whilst Flickr30k is split into 29,000 images for training, 1,014 images for validation, and 1,000 images for testing. Each image in both datasets has five different captions. 4,096-element image feature vectors that were extracted from the pre-trained VGG CNN \cite{Simonyan2014} are also available in the distributed datasets. We normalised the image vectors to unit length during preprocessing.

Tokens with frequency lower than a threshold in the training set were replaced with the `unknown' token. In our experiments we varied the threshold between 3 and 5 in order to measure the performance of each model as vocabulary size changes. For thresholds of 3, 4, and 5, this gives vocabulary sizes of 2,539, 2,918, and 3,478 for Flickr8k and 7,415, 8,275, 9,584 and  for Flickr30k.

Since our purpose is to compare the performance of architectures, we used the `barest' models possible, with the fewest number of hyperparameters. This means that complexities that are usually introduced in order to reach state-of-the-art performance, such as regularization, were avoided, since it is difficult to determine which combination of hyperparameters do not give an unfair advantage to one architecture over the other.

We constructed a basic neural language model consisting of a word embedding matrix, a basic LSTM \cite{Hochreiter1997}, and a softmax layer. The LSTM is defined as follows:
\begin{align}
i_n &= sig(x_n W_{xi} + s_{n-1} W_{si} + b_i) \\
f_n &= sig(x_n W_{xf} + s_{n-1} W_{sf} + b_f) \\
o_n &= sig(x_n W_{xo} + s_{n-1} W_{so} + b_o) \\
g_n &= \tanh(x_n W_{xc} + s_{n-1} W_{sc} + b_c) \\
c_n &= f_n \odot c_{n-1} + i_n \odot g_n \\
s_n &= o_n \odot \tanh(c_n)
\end{align}
where $x_n$ is the $n^\text{th}$ input, $s_n$ is the hidden state after $n$ inputs, $s_0$ is the all-zeros vector, $c_n$ is the cell state after $n$ inputs, $c_0$ is the all-zeros vector, $i_n$ is the input gate after $n$ inputs, $f_n$ is the forget gate after $n$ inputs, $o_n$ is the output gate after $n$ inputs, $i_n$ is the input gate after $n$ inputs, $g_n$ is the modified input used to calculate $c_n$ after $n$ inputs, $W_{\alpha \beta}$ is the weight matrix between $\alpha$ and $\beta$, $b_\alpha$ is the bias vector for $\alpha$, $\odot$ is the elementwise vector multiplication operator, and `sig' refers to the sigmoid function. The hidden state and the cell state always have the same size.

In the experiments, this basic neural language model is used as a part of two different architectures: In the inject architecture, the image vector is concatenated with each of the word vectors in a caption. In the merge architecture, it is only concatenated with the final LSTM state. The layer sizes of the embedding, LSTM state, and projected image vector were also varied in the experiments in order to measure the effect of increasing the capacity of the networks. The layer sizes used are 128, 256, and 512. The details of the architectures used in the experiments are illustrated in Figure~\ref{fig:architectures}. 

Training was performed using the Adam optimisation algorithm \cite{P.Kingma2014} with default hyperparameters and a minibatch size of 50 captions. The cost function used was sum cross-entropy. Training was carried out with an early stopping criterion which terminated training as soon as performance on the validation data started to deteriorate (validation performance is measured after each training epoch). Initialization of weights was done using Xavier initialization \cite{Glorot2010} and biases were set to zero.

Each architecture was trained three separate times; the results reported below are averages over these three separate runs.

To evaluate the trained models we generated captions for images in the test set using beam search with a beam width of 3 and a clipped maximum length of 20 words. The MSCOCO evaluation code\footnote{\url{https://github.com/tylin/coco-caption}} was used to measure the quality of the captions by using the standard evaluation metrics BLEU-(1,2,3,4) \cite{Papineni2002}, METEOR \cite{Banerjee2005}, CIDEr \cite{Vedantam2015}, and ROUGE-L \cite{Lin2004}. We also calculated the percentage of word types that were actually used in the generated captions out of the vocabulary of available word types. This measure indicates how well each architecture exploits the vocabulary it is trained on.

The code used for the experiments was implemented with TensorFlow and is available online\footnote{\url{https://github.com/mtanti/rnn-role}}.

\section{Results}\label{sec:results}

Table~\ref{tbl:results} reports means and standard deviations over the three runs of all the MSCOCO measures and the vocabulary usage. Since the point is to compare the effects of the architectures rather than to reach state-of-the-art performance, we do not include results from other published systems in our tables.

\begin{table*}[!t]
\centering
\begin{scriptsize}
\setlength\tabcolsep{2pt}

\begin{subtable}{\textwidth}
  \centering
  \begin{tabular}{|cc|cc|cc|cc|cc|}
  \hline
  & & \multicolumn{2}{|c|}{\bf \% Vocabulary} & \multicolumn{2}{|c|}{\bf CIDEr} & \multicolumn{2}{|c|}{\bf METEOR} & \multicolumn{2}{|c|}{\bf ROUGE-L} \\
  \bf Layer & \bf Vocab. & \bf Merge & Inject & \bf Merge & \bf Inject & \bf Merge & \bf Inject & \bf Merge & \bf Inject \\
  \hline
  128	& 2539	& \bf 14.730 (0.40)	& 10.555 (0.34)	& \bf 0.460 (0.01)	& 0.431 (0.01)	& \bf 0.192 (0.00)	& 0.183 (0.00)	& \bf 0.445 (0.00)	& 0.430 (0.00) \\
  128	& 2918	& \bf 13.719 (0.49)	& 8.876 (0.24)	& \bf 0.456 (0.00)	& 0.431 (0.00)	& \bf 0.191 (0.00)	& 0.185 (0.00)	& \bf 0.437 (0.00)	& 0.434 (0.00) \\
  128	& 3478	& \bf 11.223 (0.35)	& 8.175 (0.31)	& \bf 0.458 (0.01)	& 0.433 (0.01)	& \bf 0.192 (0.00)	& 0.187 (0.00)	& \bf 0.442 (0.00)	& 0.432 (0.00) \\
  \hline
  256	& 2539	& \bf 15.439 (0.84)	& 11.448 (0.71)	& \bf 0.462 (0.01)	& 0.456 (0.01)	& \bf 0.192 (0.00)	& 0.189 (0.00)	& \bf 0.439 (0.00)	& 0.436 (0.00) \\
  256	& 2918	& \bf 13.697 (0.19)	& 10.430 (0.34)	& \bf 0.456 (0.01)	& 0.451 (0.01)	& \bf 0.190 (0.00)	& 0.189 (0.00)	& 0.438 (0.00)	& \bf 0.440 (0.00) \\
  256	& 3478	& \bf 11.252 (0.51)	& 8.405 (0.39)	& \bf 0.470 (0.01)	& 0.449 (0.02)	& \bf 0.191 (0.00)	& 0.189 (0.00)	& \bf 0.439 (0.00)	& 0.437 (0.00) \\
  \hline
  512	& 2539	& \bf 15.741 (0.40)	& 12.761 (0.81)	& 0.452 (0.01)	& \bf 0.464 (0.00)	& 0.191 (0.00)	& \bf 0.192 (0.00)	& 0.437 (0.00)	& \bf 0.442 (0.00) \\
  512	& 2918	& \bf 13.114 (0.75)	& 10.155 (0.42)	& \bf 0.469 (0.01)	& 0.457 (0.00)	& \bf 0.193 (0.00)	& 0.189 (0.00)	& \bf 0.440 (0.00)	& 0.437 (0.00) \\
  512	& 3478	& \bf 11.501 (0.49)	& 8.587 (0.50)	& \bf 0.458 (0.01)	& 0.439 (0.01)	& \bf 0.192 (0.00)	& 0.188 (0.00)	& \bf 0.439 (0.00)	& 0.434 (0.00) \\
  \hline
  \end{tabular}
  \caption{\label{tbl:results_flickr8k_1} Flickr8k: \% of vocabulary used, CIDEr, METEOR and ROUGE-L results.}
\end{subtable}
\vspace{10pt}

\begin{subtable}{\textwidth}
  \centering
  \begin{tabular}{|cc|cc|cc|cc|cc|}
  \hline
  & & \multicolumn{2}{|c|}{\bf BLEU-1} & \multicolumn{2}{|c|}{\bf BLEU-2} & \multicolumn{2}{|c|}{\bf BLEU-3} & \multicolumn{2}{|c|}{\bf BLEU-4} \\
  \bf Layer & \bf Vocab. & \bf Merge & \bf Inject & \bf Merge & \bf Inject & \bf Merge & \bf Inject & \bf Merge & \bf Inject \\
  \hline
  128	& 2539	& \bf 0.600 (0.00)	& 0.592 (0.01)	& \bf 0.410 (0.00)	& 0.405 (0.01)	& \bf 0.272 (0.00)	& 0.270 (0.01)	& \bf 0.179 (0.00)	& 0.177 (0.00) \\
  128	& 2918	& \bf 0.595 (0.01)	& 0.590 (0.00)	& 0.405 (0.01)	& \bf 0.406 (0.00)	& 0.267 (0.01)	& \bf 0.271 (0.00)	& 0.175 (0.00)	& \bf 0.178 (0.00) \\
  128	& 3478	& \bf 0.608 (0.01)	& 0.586 (0.01)	& \bf 0.416 (0.01)	& 0.401 (0.01)	& \bf 0.276 (0.01)	& 0.268 (0.01)	& \bf 0.182 (0.01)	& 0.178 (0.01) \\
  \hline
  256	& 2539	& \bf 0.594 (0.00)	& 0.591 (0.00)	& 0.407 (0.01)	& \bf 0.408 (0.00)	& 0.269 (0.01)	& \bf 0.276 (0.00)	& 0.176 (0.01)	& \bf 0.184 (0.00) \\
  256	& 2918	& \bf 0.596 (0.01)	& 0.596 (0.01)	& 0.405 (0.01)	& \bf 0.413 (0.01)	& 0.265 (0.00)	& \bf 0.278 (0.01)	& 0.172 (0.00)	& \bf 0.184 (0.00) \\
  256	& 3478	& \bf 0.601 (0.00)	& 0.596 (0.01)	& \bf 0.411 (0.00)	& 0.409 (0.01)	& 0.272 (0.01)	& \bf 0.274 (0.01)	& 0.179 (0.01)	& \bf 0.181 (0.01) \\
  \hline
  512	& 2539	& 0.597 (0.01)	& \bf 0.603 (0.00)	& 0.406 (0.01)	& \bf 0.419 (0.00)	& 0.267 (0.01)	& \bf 0.283 (0.00)	& 0.176 (0.01)	& \bf 0.188 (0.00) \\
  512	& 2918	& \bf 0.593 (0.01)	& 0.589 (0.01)	& 0.404 (0.01)	& \bf 0.409 (0.00)	& 0.268 (0.00)	& \bf 0.277 (0.00)	& 0.177 (0.00)	& \bf 0.185 (0.00) \\
  512	& 3478	& \bf 0.597 (0.01)	& 0.587 (0.00)	& \bf 0.407 (0.01)	& 0.405 (0.00)	& 0.270 (0.01)	& \bf 0.272 (0.00)	& 0.178 (0.00)	& \bf 0.180 (0.01) \\
  \hline
  \end{tabular}
  \caption{\label{tbl:results_flickr8k_2} Flickr8k: BLEU-$n$ scores.}
\end{subtable}
\vspace{10pt}

\begin{subtable}{\textwidth}
  \centering
  \begin{tabular}{|cc|cc|cc|cc|cc|}
  \hline
& & \multicolumn{2}{|c|}{\bf \% Vocabulary} & \multicolumn{2}{|c|}{\bf CIDEr} & \multicolumn{2}{|c|}{\bf METEOR} & \multicolumn{2}{|c|}{\bf ROUGE-L} \\
  \bf Layer & \bf Vocab. & \bf Merge & Inject & \bf Merge & \bf Inject & \bf Merge & \bf Inject & \bf Merge & \bf Inject \\
  \hline
  128	& 7415	& \bf 6.253 (0.06)	& 5.255 (0.02)	& \bf 0.362 (0.01)	& 0.339 (0.01)	& \bf 0.174 (0.00)	& 0.169 (0.00)	& \bf 0.417 (0.00)	& 0.415 (0.00) \\
  128	& 8275	& \bf 5.402 (0.20)	& 4.939 (0.08)	& \bf 0.376 (0.00)	& 0.351 (0.00)	& \bf 0.174 (0.00)	& 0.171 (0.00)	& \bf 0.420 (0.00)	& 0.417 (0.00) \\
  128	& 9584	& \bf 4.793 (0.01)	& 4.090 (0.18)	& \bf 0.378 (0.00)	& 0.355 (0.00)	& \bf 0.175 (0.00)	& 0.171 (0.00)	& \bf 0.420 (0.00)	& 0.419 (0.00) \\
  \hline
  256	& 7415	& \bf 6.150 (0.18)	& 5.597 (0.11)	& \bf 0.363 (0.00)	& 0.361 (0.01)	& \bf 0.174 (0.00)	& 0.173 (0.00)	& 0.414 (0.00)	& \bf 0.420 (0.00) \\
  256	& 8275	& \bf 5.559 (0.08)	& 5.410 (0.10)	& \bf 0.364 (0.01)	& 0.359 (0.00)	& \bf 0.174 (0.00)	& 0.173 (0.00)	& 0.416 (0.00)	& \bf 0.417 (0.00) \\
  256	& 9584	& \bf 4.873 (0.07)	& 4.309 (0.18)	& \bf 0.364 (0.01)	& 0.359 (0.01)	& \bf 0.175 (0.00)	& 0.173 (0.00)	& 0.416 (0.00)	& \bf 0.420 (0.00) \\
  \hline
  512	& 7415	& \bf 6.330 (0.56)	& 5.732 (0.32)	& 0.365 (0.01)	& \bf 0.367 (0.01)	& \bf 0.173 (0.00)	& 0.173 (0.00)	& 0.416 (0.00)	& \bf 0.422 (0.01) \\
  512	& 8275	& \bf 5.619 (0.09)	& 5.221 (0.49)	& \bf 0.370 (0.00)	& 0.369 (0.01)	& \bf 0.174 (0.00)	& 0.174 (0.00)	& 0.419 (0.00)	& \bf 0.422 (0.00) \\
  512	& 9584	& \bf 4.887 (0.16)	& 4.309 (0.25)	& 0.357 (0.01)	& \bf 0.360 (0.01)	& 0.172 (0.00)	& \bf 0.172 (0.00)	& 0.414 (0.00)	& \bf 0.417 (0.00) \\
  \hline
  \end{tabular}
  \caption{\label{tbl:results_flickr30k_1} Flickr30k: \% of vocabulary used, CIDEr, METEOR and ROUGE-L results.}
\end{subtable}
\vspace{10pt}

\begin{subtable}{\textwidth}
  \centering
  \begin{tabular}{|cc|cc|cc|cc|cc|}
  \hline
  & & \multicolumn{2}{|c|}{\bf BLEU-1} & \multicolumn{2}{|c|}{\bf BLEU-2} & \multicolumn{2}{|c|}{\bf BLEU-3} & \multicolumn{2}{|c|}{\bf BLEU-4} \\
  \bf Layer & \bf Vocab. & \bf Merge & \bf Inject & \bf Merge & \bf Inject & \bf Merge & \bf Inject & \bf Merge & \bf Inject \\
  \hline
  128	& 7415	& \bf 0.601 (0.01)	& 0.595 (0.01)	& \bf 0.403 (0.01)	& 0.400 (0.01)	& \bf 0.268 (0.01)	& 0.265 (0.01)	& \bf 0.179 (0.01)	& 0.175 (0.01) \\
  128	& 8275	& \bf 0.605 (0.01)	& 0.604 (0.00)	& \bf 0.411 (0.01)	& 0.409 (0.00)	& \bf 0.276 (0.01)	& 0.275 (0.00)	& \bf 0.185 (0.00)	& 0.183 (0.00) \\
  128	& 9584	& \bf 0.610 (0.01)	& 0.605 (0.00)	& \bf 0.414 (0.01)	& 0.411 (0.00)	& \bf 0.278 (0.00)	& 0.275 (0.01)	& \bf 0.186 (0.00)	& 0.184 (0.01) \\
  \hline
  256	& 7415	& 0.593 (0.01)	& \bf 0.606 (0.00)	& 0.400 (0.01)	& \bf 0.412 (0.00)	& 0.268 (0.01)	& \bf 0.277 (0.00)	& 0.179 (0.01)	& \bf 0.186 (0.01) \\
  256	& 8275	& 0.594 (0.01)	& \bf 0.603 (0.01)	& 0.402 (0.01)	& \bf 0.409 (0.00)	& 0.269 (0.01)	& \bf 0.275 (0.00)	& 0.180 (0.00)	& \bf 0.183 (0.00) \\
  256	& 9584	& 0.596 (0.01)	& \bf 0.614 (0.01)	& 0.404 (0.00)	& \bf 0.419 (0.01)	& 0.270 (0.00)	& \bf 0.283 (0.00)	& 0.181 (0.00)	& \bf 0.189 (0.00) \\
  \hline
  512	& 7415	& 0.598 (0.02)	& \bf 0.617 (0.01)	& 0.404 (0.02)	& \bf 0.422 (0.01)	& 0.270 (0.01)	& \bf 0.285 (0.00)	& 0.181 (0.01)	& \bf 0.191 (0.00) \\
  512	& 8275	& 0.603 (0.00)	& \bf 0.609 (0.01)	& 0.406 (0.00)	& \bf 0.419 (0.01)	& 0.271 (0.00)	& \bf 0.284 (0.01)	& 0.181 (0.00)	& \bf 0.191 (0.00) \\
  512	& 9584	& 0.596 (0.00)	& \bf 0.609 (0.01)	& 0.399 (0.00)	& \bf 0.414 (0.01)	& 0.265 (0.00)	& \bf 0.278 (0.01)	& 0.177 (0.00)	& \bf 0.185 (0.00) \\
  \hline
  \end{tabular}
  \caption{\label{tbl:results_flickr30k_2} Flickr30k: BLEU-$n$ scores.}
\end{subtable}
\vspace{10pt}

\end{scriptsize}
\caption{\label{tbl:results} Results on the captions generated using the inject and merge architectures. Values are means over three separately retrained models, together with the standard deviation in parentheses. Legend: Layer - the layer size used (`x' in Figure~\ref{fig:architectures}); Vocab. - the vocabulary size used.}
\end{table*}

Across all experimental variables (dataset, vocabulary, and layer sizes), the performance of the merge architecture is generally superior to that of the inject architecture in all measures except for ROUGE-L and BLEU (ROUGE-L is designed for evaluating text summarization whilst BLEU is criticized for its lack of correlation with human-given scores). In what follows, we focus on the CIDEr measure for caption quality as it was specifically designed for captioning systems.

Although merge outperforms inject by a rather narrow margin, the low standard deviation over the three training runs suggests that this is a consistent performance advantage across train-and-test runs. In any case, there is clearly no disadvantage to the merge strategy with respect to injecting image features.

One peculiarity is that results on Flickr8k are better than those on Flickr30k. This could mean that Flickr8k captions contain less variation, hence are easier to perform well on. Preliminary results on the larger dataset MSCOCO \cite{Lin2014} (currently in progress) show CIDEr results over 0.7 which means that either Flickr8k is too easy or Flickr30k is too hard when compared to the much larger MSCOCO.

The best-performing models are merge with state size of 256 on Flickr8k, and merge with state size 128 on Flickr30k, both with minimum token frequency threshold of 3. Inject models tend to improve with increasing state size, on both datasets, while the relationship between the performance of merge and the state size shows no discernible trend. Inject therefore does not seem to overfit as state size increases, even on the larger dataset. At the same time, inject only seems to be able to outperform the best scores achieved by merge if it has a much larger layer size. Therefore, in practical terms, inject models have to have larger capacity to be at par with merge. Put differently, merge has a higher performance to model size ratio and makes more efficient use of limited resources (this observation holds even when model size is defined in terms of number of parameters instead of layer sizes).

Given the same layer sizes and vocabulary, the number of parameters for merge is greater than for inject. The difference becomes greater as the vocabulary size is increased. For a vocabulary size of 2,539 and layer size of 512, merge has about 3\% more parameters than inject whilst for a vocabulary size of 9,584 and layer size of 512, merge has about 20\% more parameters. However, the foregoing remarks concerning over- and under-fitting also apply when the difference between the number of parameters is small. That is, the difference in performance is due at least in part to architectural differences, not just to differences in number of parameters.

Merge models use a greater proportion of the training vocabulary on test captions. However, the proportion of vocabulary used is generally quite small for both architectures: less than 16\% for Flickr8k and less than 7\% for Flickr30k. Overall, the trend is for smaller proportions of the overall training vocabulary to be used, as the vocabulary grows larger, suggesting that neural language models find it harder to use infrequent words (which are more numerous at larger vocabulary sizes, by definition). In practice, it means that reducing training vocabularies results in minimal performance loss.

Overall, the evidence suggests that delaying the merging of image features with linguistic encodings to a late stage in the architecture may be advantageous, at least as far as corpus-based evaluation measures are concerned. Furthermore, the results suggest that a merge architecture has a higher capacity than an inject architecture and can generate better quality captions with smaller layers.

\section{Discussion}

If the RNN had the primary role of generating captions, then it would need to have access to the image in order to know what to generate. This does not seem to be the case as including the image into the RNN is not generally beneficial to its performance as a caption generator.

When viewing RNNs as having the primary role of encoding rather than generating, it makes sense that the inject architecture generally suffers in performance when compared to the merge architecture. The most plausible explanation has to do with the handling of variation. Consider once more the task of the RNN in the image captioning task: During training, captions are broken down into prefixes of increasing length, with each prefix compressed to a fixed-size vector, as illustrated in Figure \ref{fig:rnn-cont} above. 

In the inject architecture, the encoding task is made more complex by the inclusion of image features. Indeed, in the version of inject used in our experiments -- the most commonly used solution in the caption generation literature\footnote{We are referring to architectures that inject image features in parallel with word embeddings in the RNN. In the literature, when this type of architecture is used, the image features might only be included with some of the words or are changed for different words (such as in attention models).} -- image features are concatenated with every word in the caption. The upshot is (a) a requirement to compress caption prefixes together with image data into a fixed-size vector and (b) a substantial growth in the vocabulary size the RNN has to handle, because each image+word is treated as a single `word'. This problem is alleviated in merge, where the RNN encodes linguistic histories only, at the expense of more parameters in the softmax layer.

One practical consequence of these findings is that, while merge models can handle more variety with smaller layers, increasing the state size of the RNN in the merge architecture is potentially quite profitable, as the entire state will be used to remember a greater variety of previously generated words. By contrast, in the inject architecture, this increase in memory would be used to better accommodate information from two distinct, but combined, modalities.

\section{Conclusions}

This paper has presented two views of the role of the RNN in an image caption generator. In the first, an RNN decides on which word is the most likely to be generated next, given what has been generated before. In multimodal generation, this view encourages architectures where the image is incorporated into the RNN along with the words that were generated in order to allow the RNN to make visually-informed predictions. 

The second view is that the RNN's role is purely memory-based and is only there to encode the sequence of words that have been generated thus far. This representation informs caption prediction at a later layer of the network as a function of both the RNN encoding and perceptual features.
This view encourages architectures where vision and langauge are brought together late, in a multimodal layer.

Caption generation turns out to perform worse, in general, when image features are injected into the RNN. Thus, the role of the RNN is better conceived in terms of the learning of linguistic representations, to be used to inform later layers in the neural network, where predictions are made based on what has been generated in the past together with the image that is guiding the generation. Had the RNN been the component primarily involved in generating the caption, it would need to be informed about the image in order to know what needs to be generated; however this line of reasoning seems to hurt performance when applied to an architecture. This suggests that it is not the case that the RNN is the main component of the caption generator that is involved in generation.

In short, given a neural network architecture that is expected to process input sequences from multiple modalities, arriving at a joint representation, it would be better to have a separate component to encode each input, bringing them together at a late stage, rather than to pass them all into the same RNN through separate input channels. With respect to the question of how language should be grounded in perceptual data, the tentative answer offered by these experiments is that the link between the symbolic and perceptual should be established late, once encoding has been performed. To this end, recurrent networks are best viewed as learning representations, not as generating sequences.

\subsection{Future work}\label{sec:future}
The experiments reported here were conducted on two separate datasets. One concern is that results on Flickr8k and Flickr30k are not entirely consistent, though the superiority of merge over inject is clear in both. We are currently extending our experiments to the larger MSCOCO dataset \cite{Lin2014}.

The insights discussed in this paper invite future research on how generally applicable the merge architecture is in different domains. We would like to investigate whether similar changes in architecture would work in sequence-to-sequence tasks such as machine translation, where instead of conditioning a language model on an image we are conditioning a target language model on sentences in a source language. A similar question arises in image processing. If a CNN were conditioned to be more sensitive to certain types of objects or saliency differences among regions of a complex image, should the conditioning vector be incorporated at the beginning, thereby conditioning the entire CNN, or would it be better to instead incorporate it in a final layer, where saliency differences would then be based on high-level visual features?

There are also more practical advantages to merge architectures, such as for transfer learning. Since merge keeps the image separate from the RNN, the RNN used for captioning can conceivably be transferred from a neural language model that has been trained on general text. This cannot be done with an inject architecture since the RNN would need to be trained to combine image and text in the input. In future work, we intend to see how the performance of a caption generator is affected when the weights of the RNN are initialized from those of a general neural language model, along lines explored in neural machine translation \cite{Ramachandran2016}.

\section*{Acknowledgments}
This work was partially funded by the Endeavour Scholarship Scheme (Malta), part-financed by the European Social Fund (ESF).

\bibliography{bibliography}
\bibliographystyle{naaclhlt2016}

\end{document}